\documentclass{article}

\usepackage{amsmath,amsfonts}
\usepackage{physics}
\usepackage{spconf,graphicx}
\usepackage{booktabs}
\usepackage{makecell}
\usepackage{subcaption} 
\usepackage{overpic}  
\usepackage{lipsum}
\usepackage{hyperref}
\usepackage[nameinlink]{cleveref}
\usepackage{subcaption}
\usepackage{dblfloatfix} 
\usepackage{placeins} 
\newcommand{\bw}{{{\bf w}}}
\newcommand{\bx}{{{\bf x}}}
\crefformat{figure}{#2\templabel~#1#3}

\title{Neuro-Inspired Deep Neural Networks with Sparse, Strong Activations}

\name{Metehan Cekic$^*$ \qquad Can Bakiskan$^*$\thanks{$^*$ Equal contribution. Code can be found at: \cite{repo}. This work was supported in part by the Army Research Office under grant W911NF-19-1-0053, and by the National Science Foundation under grants CIF-1909320 and CIF-2224263.} \qquad Upamanyu Madhow}

\address {
  Department of Electrical and Computer Engineering\\
  University of California Santa Barbara, Santa Barbara, CA 93106 \\
  \texttt{\{metehancekic,canbakiskan,madhow\}@ucsb.edu} \\
}

\begin{document}
\maketitle
\begin{abstract}

While end-to-end training of Deep Neural Networks (DNNs) yields state of the art performance in an increasing array of applications, it does not provide insight into, or control over, the features being extracted.  We report here on a promising neuro-inspired approach to DNNs with sparser and stronger activations. We use standard stochastic gradient training, supplementing the end-to-end discriminative cost function with layer-wise costs promoting Hebbian (“fire together,” “wire together”) updates for highly active neurons, and anti-Hebbian updates for the remaining neurons.  Instead of batch norm, we use divisive normalization of activations (suppressing weak outputs using strong outputs),
along with implicit $\ell_2$  normalization of neuronal weights.  Experiments with standard image classification tasks on CIFAR-10 demonstrate that, relative to baseline end-to-end trained architectures, our proposed architecture (a) leads to sparser activations (with only a slight compromise on accuracy), (b) exhibits more robustness to noise (without being trained on noisy data), (c) exhibits more robustness to adversarial perturbations (without adversarial training).

\end{abstract}


\vspace*{-0.4cm}
\section{Introduction}
\label{sec:intro}

\vspace*{-0.2cm}

Since their original breakthrough in image classification performance, DNNs trained with backpropagation have attained outstanding performance in
a wide variety of fields \cite{brown2020language,silver2018general,akkaya2019solving, senior2020improved}. Yet there remain fundamental concerns
regarding their lack of interpretability and robustness (e.g, to noise, distribution shifts, and adversarial perturbations). In this paper, we explore the thesis that a first
step to alleviating these problems is to exert more control on the features being extracted by DNNs. Specifically, while standard DNNs produce a large fraction of small
activations at each layer, we seek architectures which produce a small fraction of strong activations, while continuing to utilize existing network architectures
for feedforward inference and existing software infrastructure for stochastic gradient training.

\subsection{Approach and Contributions}
\vspace*{-0.1cm}

In order to attain sparse, strong activations at each layer, we employ the following neuro-inspired strategy for modifying standard DNN training and architecture:\\
{\it Hebbian/anti-Hebbian (HaH) Training:} We supplement a standard end-to-end discriminative cost function with layer-wise costs at each layer which promote neurons producing large activations and demote neurons producing smaller activations.  The goal is to develop a neuronal basis that produces a distributed sparse code, without requiring a reconstruction
cost as in standard sparse coding \cite{olshausen1997sparse}.\\
{\it Neuronal Competition via Normalization:} We further increase sparsity by introducing Divisive Normalization (DN), which enables larger activations to suppress smaller activations.
In order to maintain a fair competition among neurons, we introduce Implicit $\ell_2$ Normalization of the neuronal weights, so that each activation may be viewed as a geometric projection of the layer input onto the ``direction'' of the neuron.  (Using implicit rather than explicit weight normalization in our inference architecture simplifies training.)

We report on experiments with CIFAR-10 image classification, comparing a baseline VGG-16 network trained end-to-end against the same architecture with HaH training and DN.
Both architectures employ implicit weight normalization, which we have verified does not adversely impact accuracy.  We demonstrate that the activations in our proposed architecture
are indeed more sparse than for the baseline network. Furthermore, robustness against noise and adversarial perturbations is enhanced, without having used noise augmentation
or adversarial training.

\begin{figure*}[!thb]
\includegraphics[width=1\linewidth,trim={0 0 0 0.2cm},clip]{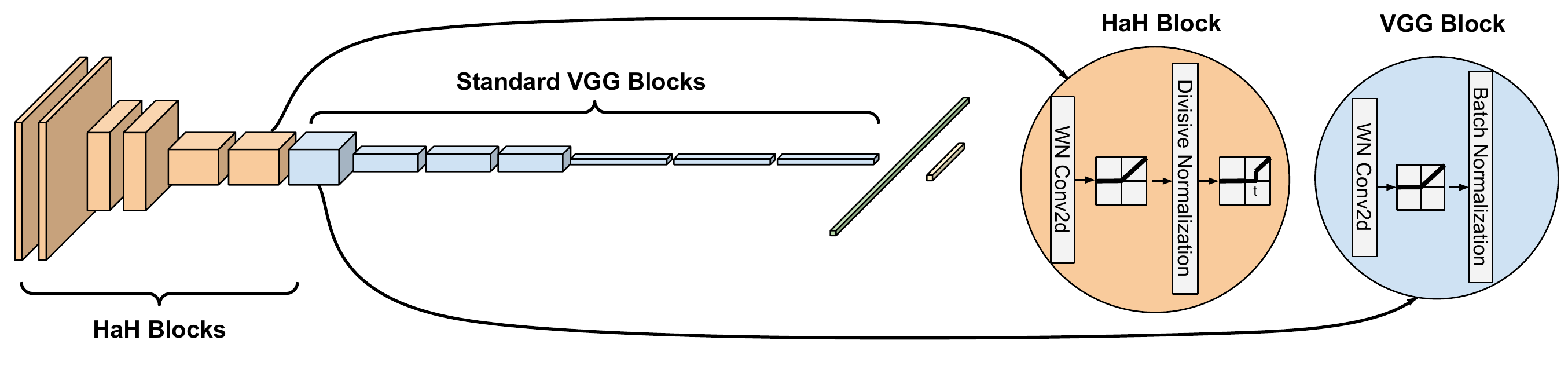}
\vspace*{-0.7cm}  
\caption{Our model consists of two different types of blocks: first 6 blocks are Hebbian-anti-Hebbian (HaH) while the rest are regular VGG blocks. HaH blocks use a weight normalized convolutional layer, followed by ReLU, divisive normalization and thresholding. Regular VGG blocks use a weight normalized convolutional layer followed by ReLU and batch norm.}   
\label{fig:model}   
\vspace*{-0.2cm}
\end{figure*}

\vspace*{-0.3cm}
\subsection{Related Work}
\vspace*{-0.1cm}

Hebbian learning has a rich history in artificial neural networks, dating back to the neocognitron \cite{Fukushima1983_neocognitron}, and including recent attempts at introducing it into deep architectures \cite{Amato2019}. However, to the best of our knowledge, ours is the first paper to clearly demonstrate gains in robustness from its incorporation in DNNs. Divisive normalization is a widely accepted concept in neuroscience \cite{carandini2012normalization,burg2021learning}, and versions of it have been shown to be competitive with other normalization techniques in deep networks  \cite{ren2016normalizing}.  Our novel contribution is in showing that divisive normalization can be engineered to enhance sparsity and robustness.  Finally, sparse coding with a reconstruction objective was shown to lead to neuro-plausible outcomes in a groundbreaking paper decades ago \cite{olshausen1997sparse}. In contrast to the iterative sparse coding and dictionary learning in such an approach, our HaH-based training targets strong sparse activations in a manner amenable to standard stochastic gradient training. 

Recent work showing potential robustness gains by directly including known aspects of mammalian vision in DNNs includes \cite{Dapello2020}, which employs Gabor filter blocks and stochasticity, and \cite{li2019learning}, which employs neural activity measurements from mice for regularization in DNNs. Rather than incorporating specific features from biological vision, we use neuro-inspiration to extract broad principles that can be folded into data-driven learning and inference in DNNs.


\vspace*{-0.4cm}
\section{Model}
\vspace*{-0.2cm}
\label{sec:hebb}

We now describe how we incorporate HaH training and divisive normalization into a standard CNN for image classification. 
We consider a ``classical'' CNN for our experiments--VGG-16 \cite{simonyan2014very} applied to
CIFAR-10, rather than variants of ResNet \cite{he2015delving}, because residual connections complicate our interpretation of building models from the bottom-up using HaH learning.  Since we wish to build robustness from the bottom up, we modify the first few convolutional blocks to incorporate neuro-inspired principles.  
We term these modified blocks ``HaH blocks.''

Each HaH block employs convolution with implicit weight normalization, followed by ReLU, then divisive normalization, and then thresholding.
Implicit weight normalization enables us to interpret the convolution outputs for each filter as projections, and we have verified that employing it in
all blocks of a baseline VGG-16 architecture does not adversely impact accuracy (indeed, it slightly improves it).  
Each standard (non-HaH) block in our architecture therefore also employs convolution with implicit weight normalization, followed by ReLU, but uses
batch norm rather than divisive normalization. Each HaH block contributes a HaH cost for training, so that the overall cost function used for training
is the standard discriminative cost and the sum of the HaH costs from the HaH blocks.

We now describe the key components of our architecture, shown in \autoref{fig:model}. 

\vspace*{-0.3cm}
\subsection{Inference in a HaH block}
\vspace*{-0.1cm}

\noindent
{\bf Implicit weight normalization:} Representing the convolution output at a given spatial location from a given filter as a tensor inner product $\langle \cdot , \cdot \rangle$ between the filter weights $\bw$ and the input $\bx$, the output of the ReLU unit following the filter is given by
\begin{equation} \label{implicit_norm}
y =  \text{ReLU} \left( \frac{\langle \bw , \bx \rangle}{|| \bw ||_2} \right)
\end{equation}
This effectively normalizes the weight tensor of each filter to unit $\ell_2$ norm, without actually having to enforce an $\ell_2$ norm constraint in the cost.  

\renewcommand{\thefigure}{3}
\begin{figure*}[!b]
\centering
\begin{minipage}{.48\textwidth}
  \centering
  \begin{overpic}[width=1\linewidth]{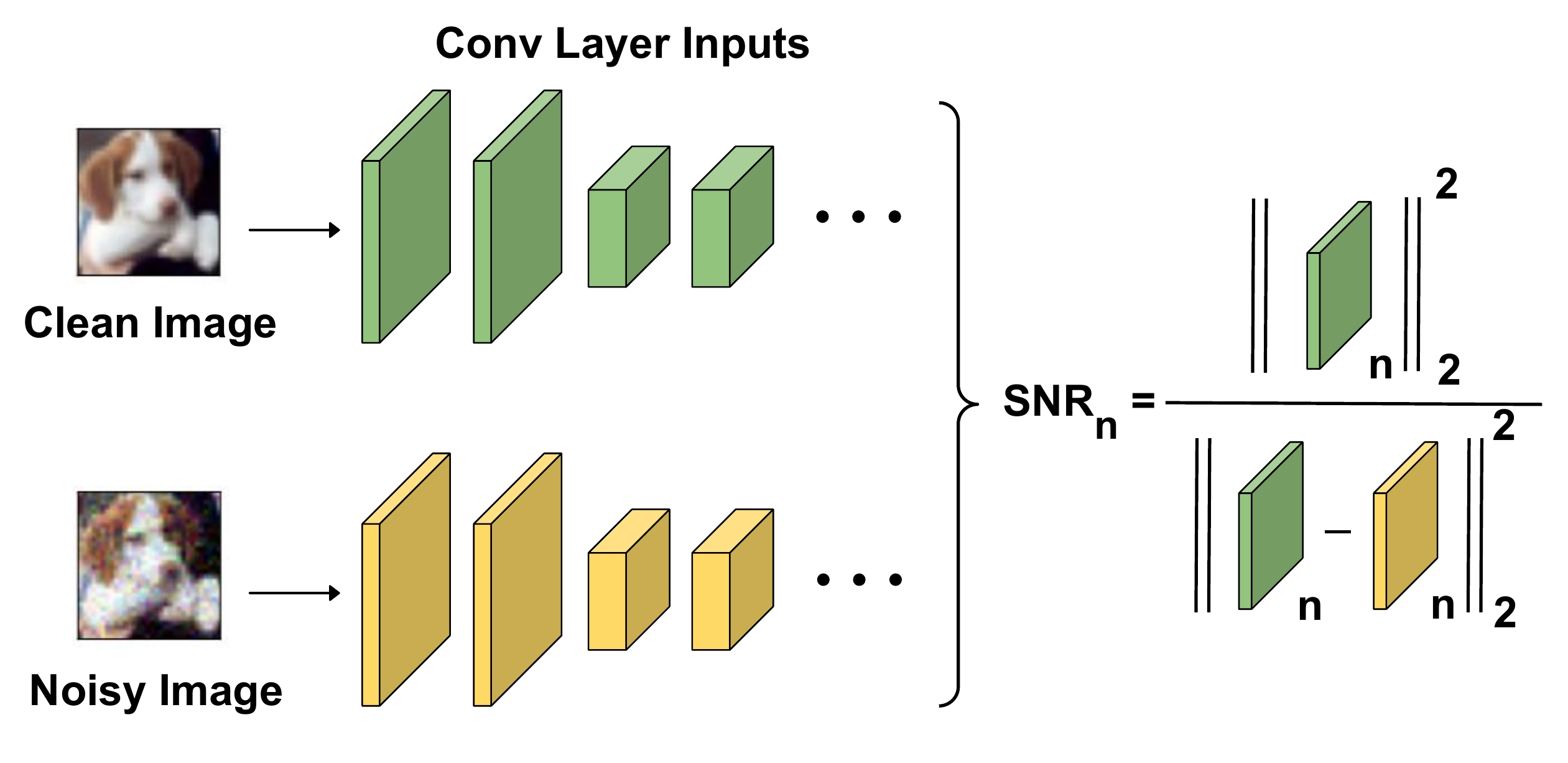}
        \put(5,49){a}
    \end{overpic}
    \phantomsubcaption
    \label{fig:snr_desc}
    
\end{minipage}%
\begin{minipage}{.52\textwidth}
  \centering
  
  \begin{overpic}[width=1\linewidth]{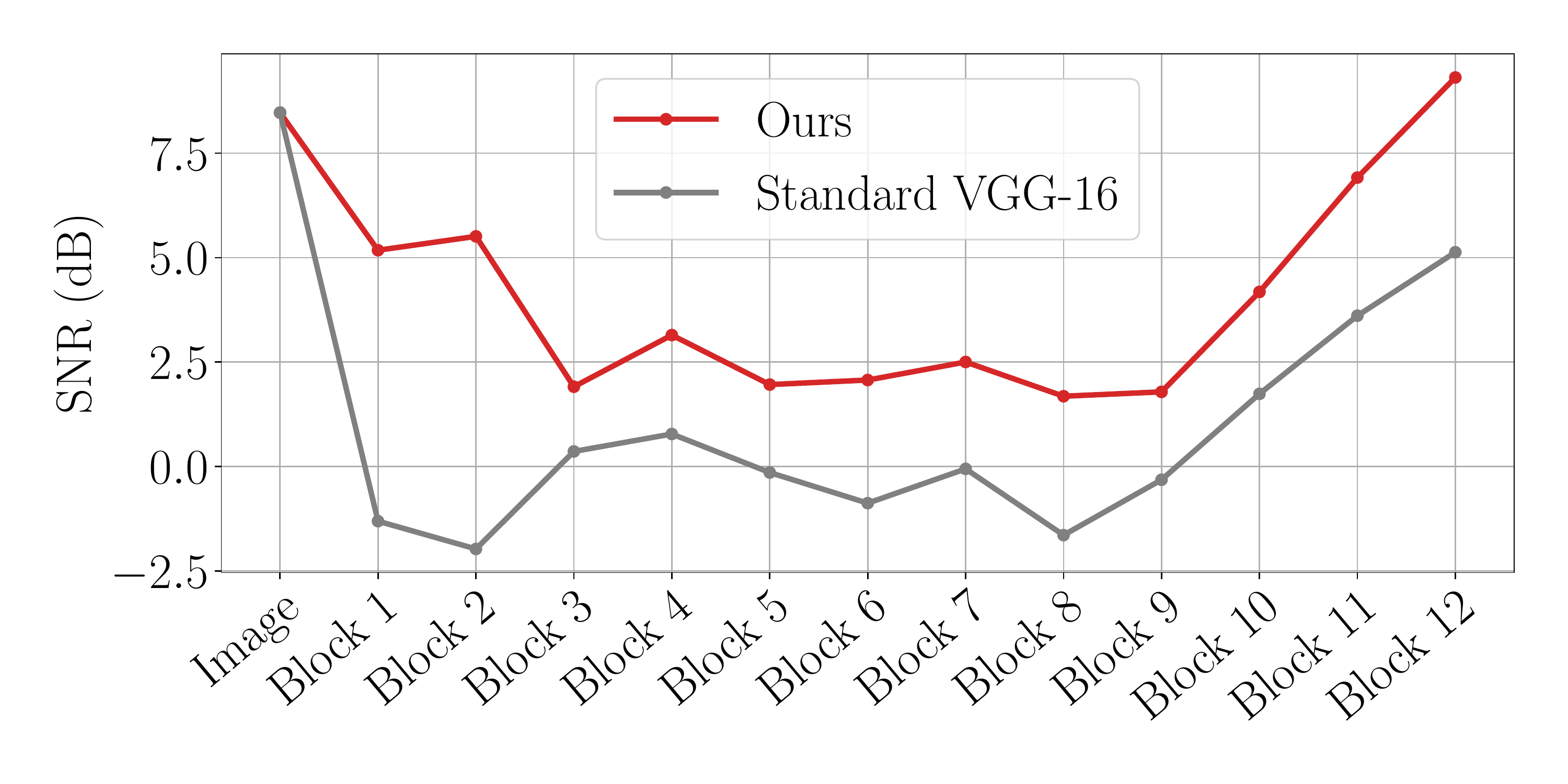}
        \put(5,47){b}
    \end{overpic}
    \phantomsubcaption
    \label{fig:snr_plot}
    
\end{minipage}   
\vspace*{-0.3cm}
\caption{\textbf{a}: To compute the SNR at the $n^{th}$ block inputs, we divide the $\ell_2$ norm of the block input corresponding to clean image by the $\ell_2$ norm of the difference of block corresponding to clean and noisy images. \textbf{b}: Comparison of SNR values of the block inputs for the standard base model (gray) and ours (red).}   
\label{fig:snr_desc_plot}
\vspace*{-0.3cm}
\end{figure*}

\noindent
{\bf Divisive normalization:} If we have $N$ filters in a given HaH block, let $y_1 (loc) ,...,y_N (loc)$ denote the corresponding activations computed as in (\autoref{implicit_norm}) for a given spatial location $loc$. Let $M(loc) = \frac{1}{N}\sum_{k=1}^N y_k (loc)$ denote the mean of the activations at a given location,
and let $M_{max} = {\rm max}_{loc} M(loc)$ denote the maximum of this mean over all locations.  We normalize each activation using these terms as follows:
\begin{equation} \label{divisive_norm}
z_k (loc) = \frac{y_k (loc)}{\sigma M_{max} + (1-\sigma ) M (loc)} ~,~k=1,...,N
\end{equation}
where $0 \leq \sigma \leq 1$ is a hyperparameter which can be separately tuned for each HaH block.
Thus, in addition to creating competition among neurons at a given location by dividing by $M (loc)$, we also include $M_{max}$ in the denominator in order
to suppress contributions at locations for which the input is ``noise'' rather than a strong enough ``signal'' well-aligned with one or more of the filters. 
This particular implementation of divisive normalization ensures that the output of a HaH-block is scale-invariant (i.e., we get the same output if we scale the input to the block
by any positive scalar). 

\noindent
{\bf Adaptive Thresholding:} Finally, we ensure that each neuron is producing significant outputs by neuron-specific thresholding after divisive normalization.
The output of the $k$th neuron at location $loc$ is given by
\begin{equation} \label{adaptive_threshold}
o_k (loc) = \biggl\{
    \begin{array}{ll}
    z_k (loc) & \text { if } z_{k} (loc) \geq \tau_k\\
    0, & \text{otherwise}
    \end{array}
\end{equation}
where the threshold $\tau_k$ is neuron and image specific.  For example, we may set $\tau_k$ to the 90th percentile of the statistics of $z_k (loc)$ in order to get an activation rate of 10\% for each neuron for every image.  Another simple choice that works well, but gives higher activation
rates, is to set $\tau_k$ to the mean of $z_k (loc)$ for each image.

\vspace*{-0.3cm}
\subsection{HaH Training}
\vspace*{-0.1cm}

For an $N$-neuron HaH block with activations $y_k (loc)$, $k=1,...,N$ at location $loc$, the Hebbian/anti-Hebbian cost seeks to maximize the average of the top $K$ activations, and to minimize the average of
the remaining $N-K$ activations, where $K$ is a hyperparameter. Thus, sorting the activations $\{ y_k (loc) \}$ so that $y^{(1)} (loc) \geq y^{(2)} (loc) \geq ... \geq y^{(N)} (loc)$, the contribution to the HaH cost (to be maximized) is given by
\begin{equation} \label{HaH_cost}
L_{block} ( loc) = \frac{1}{K} \sum_{k=1}^K y^{(k)} (loc) -  \lambda  \frac{1}{N-K} \sum_{k=K+1}^N y^{(k)} (loc)
\end{equation}
where $\lambda \geq 0$ is a hyperparameter determining how much to emphasize the anti-Hebbian component of the adaptation. 
The overall HaH cost for the block, $L_{block}$, which we wish to maximize, is simply the mean over all locations and images.

The overall loss function to be minimized is now given by
\begin{equation} \label{loss_function}
L = L_{disc} - \sum_{{\rm HaH~blocks}} \alpha_{block} L_{block} 
\end{equation}
where $L_{disc}$ is the standard discriminative loss, and $\{ \alpha_{block} \geq 0 \}$ are hyper-parameters determining the relative weight of the HaH costs across blocks.  

\renewcommand{\thefigure}{2}
\begin{figure}[!t]
  \centering
    \vspace*{-0.3cm}
    \hspace*{-0.2cm}
  \includegraphics[width=1.02\linewidth,trim={0.5cm 0.5cm 0.5cm 0.9cm},clip]{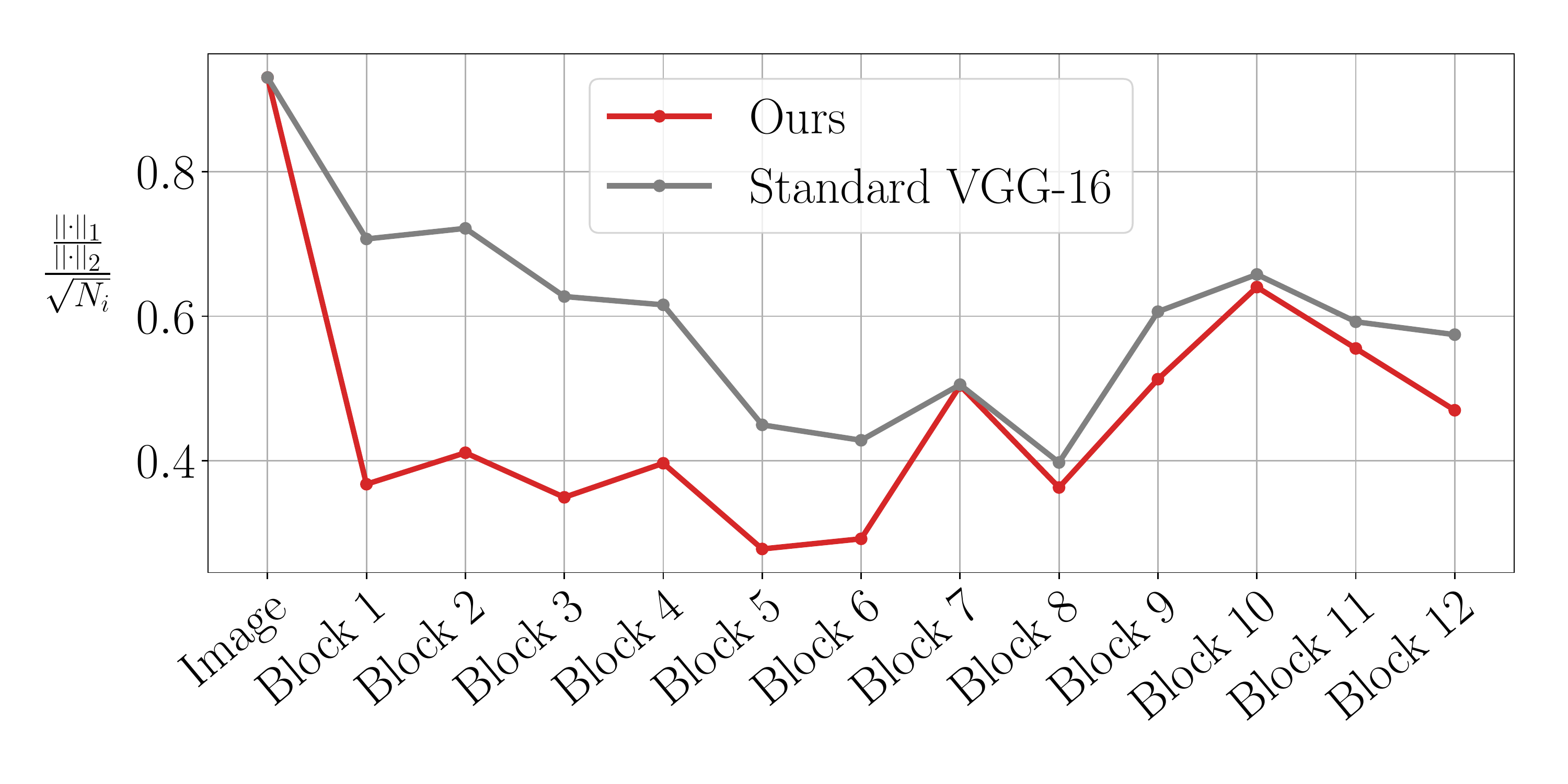}
  \vspace*{-0.7cm} 
\caption{HaH blocks yield sparser activations than baseline. The measure of sparsity is the Hoyer ratio  \cite{hoyer2004non} of $\ell_1$ norm to  $\ell_2$ norm of activations across channels, averaged across spatial locations, and then normalized to lie in [0,1] (lower values correspond to more sparsity).}   
   \label{fig:sparsity} 
   \vspace*{-0.4cm} 
\end{figure}

\vspace*{-0.3cm}
\section{Experiments}
\vspace*{-0.2cm}
\label{sec:exp}

We consider VGG-16 with the first 6 blocks (each block includes conv, ReLU, batch norm) replaced by HaH blocks (each block includes conv, ReLU, divisive norm, thresholding). In our training, we use  Adam optimizer \cite{kingma2014adam} with an initial learning rate of $10^{-3}$, multiplied by 0.1 at epoch 60 and again at epoch 80. We train all models for 100 epochs on CIFAR-10. We choose $\tau_k$ in \autoref{adaptive_threshold} to keep 20\% of activations. We use $[4.5\times 10^{-3},\: 2.5\times 10^{-3},\: 1.3\times 10^{-3},\: 1\times 10^{-3},\: 8\times 10^{-4},\: 5\times 10^{-4}]$ for $\alpha$ in \autoref{loss_function}. We use $0.1$ for $\lambda$ and set $K$ to 10\% of number of filters in each layer in \autoref{HaH_cost} and set $\sigma=0.1$ in \autoref{divisive_norm}. Details about other hyper-parameters can be found in our code  repository \cite{repo}.

\noindent
{\bf Sparser activations:} \autoref{fig:sparsity} shows that the activations in these first 6 blocks are indeed more sparse for our architecture than for baseline VGG.

\noindent
{\bf Enhanced robustness to noise:} We borrow the concept of signal-to-noise-ratio (SNR) from wireless communication to obtain a block-wise measure of robustness. Let $f_n (x )$ denote the input tensor at block $n$ in response to clean image $x$, and $f_n(x + w )$ the input tensor when the image is corrupted by noise $w$.  As illustrated in \autoref{fig:snr_desc}, we define SNR as
\vspace*{-0.1cm}
\begin{equation}
\text{SNR}_n = 10\log_{10}\left(\mathbb{E}_{x\sim \mathcal{D}_{test}}\left[\frac{||f_n(x)||_2^2}{||f_n(x+w) - f_n(x)||_2^2}\right]\right) dB
\label{eq:snr}
\end{equation}
converting to logarithmic decibel (dB) scale as is common practice. \autoref{fig:snr_plot} shows that the SNR for our model comfortably exceeds that of the standard model, especially in the first 6 HaH blocks.





These higher SNR values also translate to gains in accuracy with noisy images:
\autoref{fig:accuracy_v_noise} compares the accuracy of our model and the base model for different levels of Gaussian noise. There are substantial accuracy gains at high noise levels: 64\% vs. 26\% at a noise standard deviation of 0.1, for example.




\setcounter{figure}{3}
\renewcommand{\thefigure}{\arabic{figure}}
\begin{figure}[!t]
\centering
\hspace*{-0.4cm}
\begin{subfigure}[b]{1.05\linewidth}
  \centering
  \includegraphics[width=1\linewidth,trim={0 0.5cm 0 0.9cm},clip]{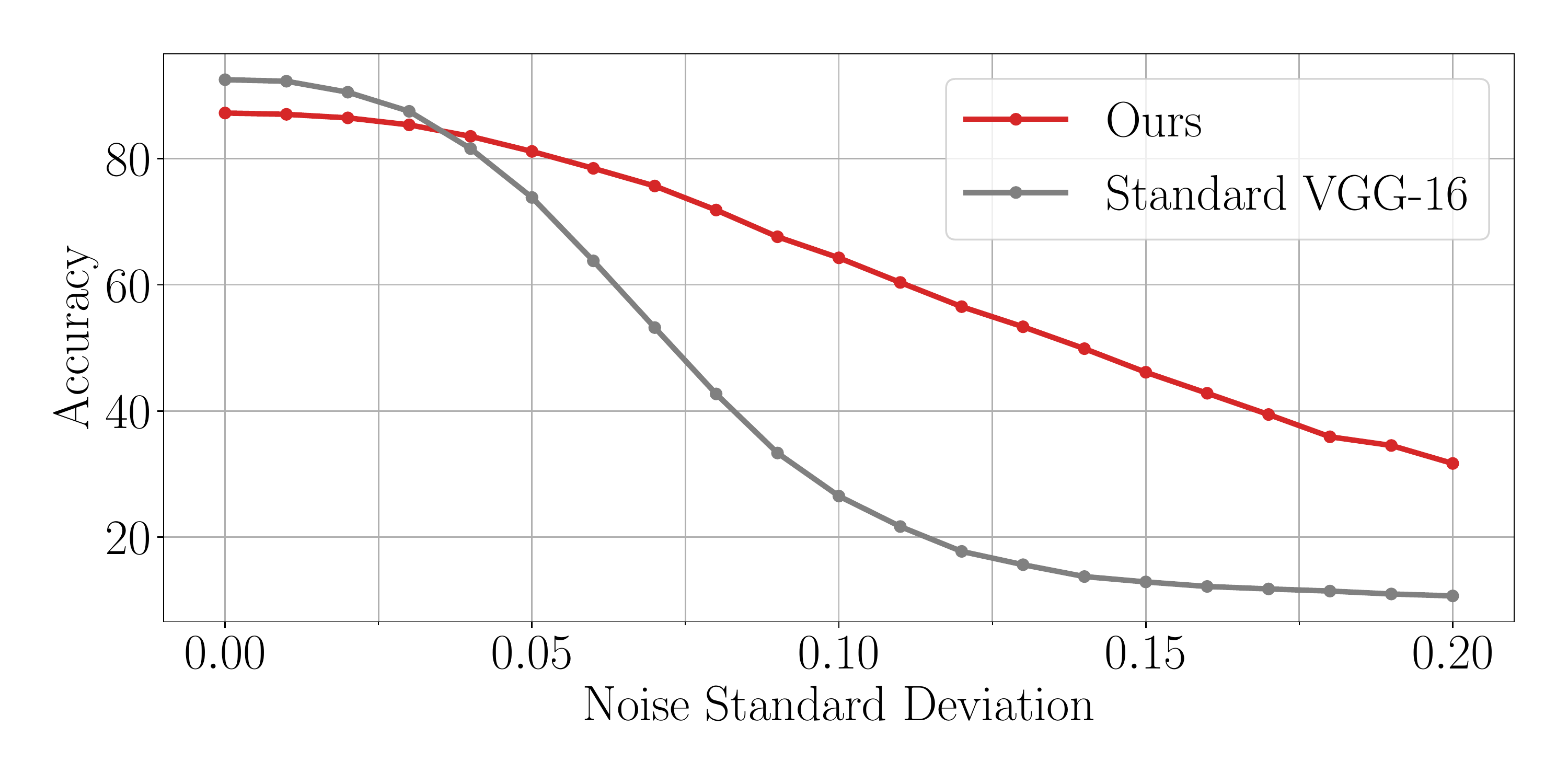}
  \vspace*{-0.3cm} 
\end{subfigure}
\hspace*{0.01cm} 
\begin{subfigure}[b]{1.035\linewidth}
  \centering
   \includegraphics[width=0.92\linewidth,trim={0 8.5cm 0 1.35cm},clip]{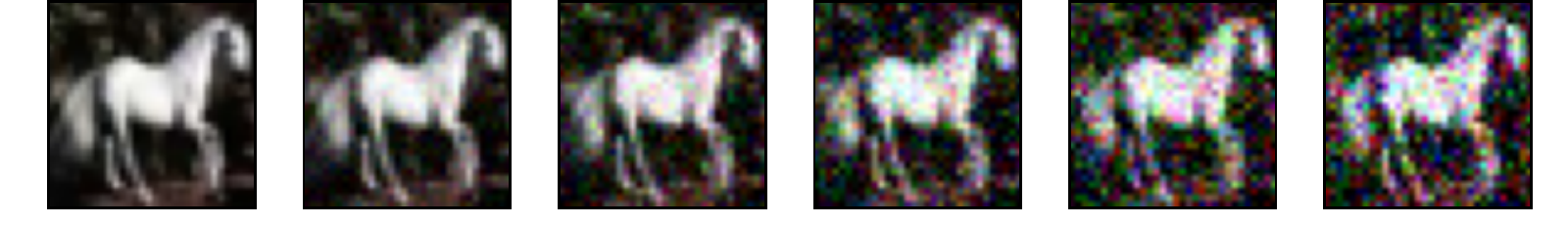}
   \label{fig:noisy_imgs}
\end{subfigure}
\caption{Comparison of classification accuracies as a function of noise $\sigma$. To provide a concrete sense of the impact of noise, noisy images  at increasing values of $\sigma$ are shown below the graph.}
\label{fig:accuracy_v_noise}    
\vspace*{-0.4cm}
\end{figure}

\begin{table}[!b]\centering
  \def\arraystretch{1.25}%
\vspace*{-0.3cm}
  \caption{Enhanced accuracy against noise and adversarial attacks}
\vspace*{-0.2cm}
  \begin{small}
  \begin{tabular}{rcccc}
    \toprule

    & Clean  &  \makecell[c]{Noisy \\ ($\sigma=0.1$)} &  \makecell[c]{Adv ($\ell_\infty$) \\ ($\epsilon=2/255$)}  &  \makecell[c]{Adv  ($\ell_2$)\\ ($\epsilon=0.25$)}   \\  \midrule
    Standard & \textbf{92.5\%} & 26.6\% & 10.4\% & 13.9\% \\
    Ours &87.3\%& \textbf{64.0\%}& \textbf{21.5\%} & \textbf{27.6\%}\\
    \bottomrule
  \end{tabular}
  \end{small}
  \label{table:standard_table}
\end{table}
\noindent
{\bf Enhanced robustness to adversarial attacks:} While we have not trained with adversarial examples, we find that, as expected, the noise rejection capabilities of the HaH blocks also translates into gains in adversarial robustness relative to the baseline VGG model. 
This holds for state-of-the-art gradient-based attacks \cite{madry2017towards,pintor2021fast}, as well as AutoAttack, an ensemble of parameter-free attacks suggested by RobustBench \cite{croce2020robustbench}. We observe no additional benefit of using gradient-free attacks, and conclude that the robustness provided by our scheme is not because of gradient-masking. Because of space constraints, we only report on results from minimum-norm adversarial attacks and AutoAttack. 

\begin{figure}[!t]
\centering
\hspace*{-0.5cm}
\includegraphics[width=1.08\linewidth,trim={0 0.2cm 0 0.2cm},clip]{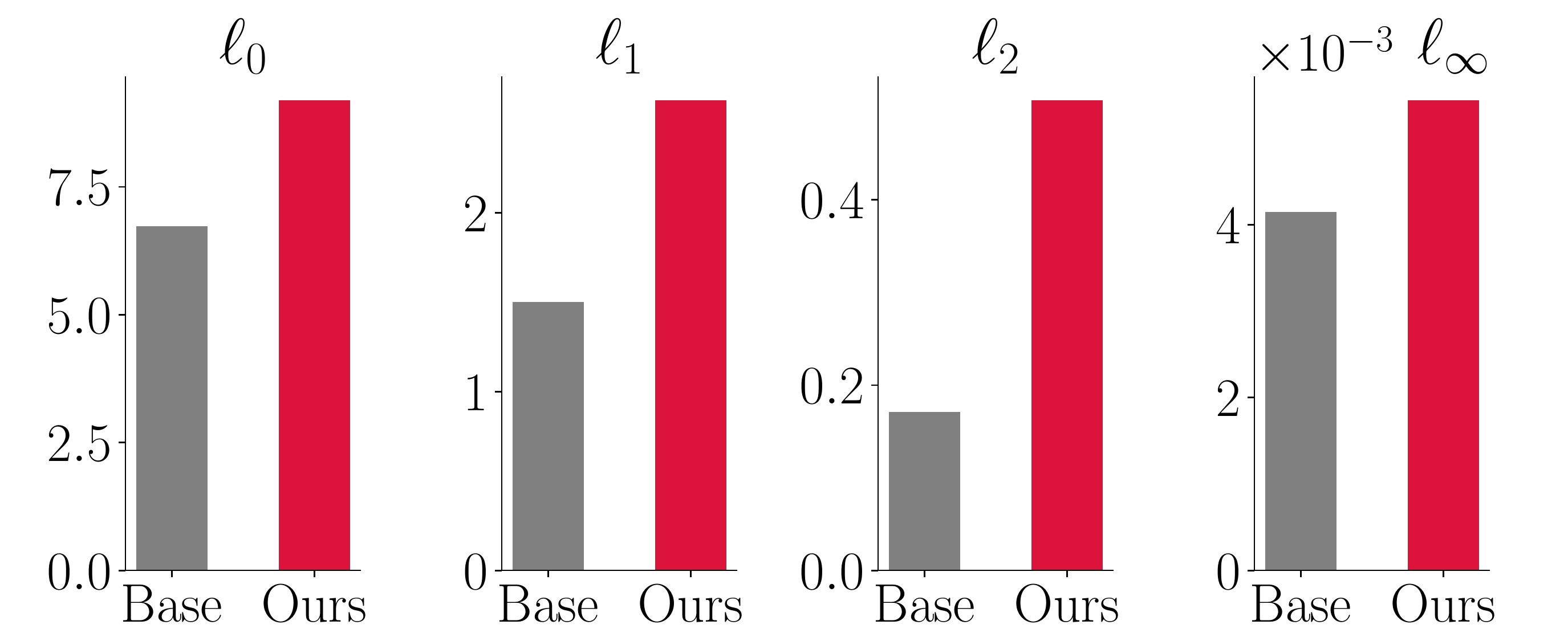}
\caption{The average norm of minimum-norm adversarial attacks is higher for our model for all $\ell_p$ norms considered.}
\label{fig:minnorm}   
\vspace*{-0.3cm}
\end{figure}

\autoref{fig:minnorm} shows that the minimum distortion needed to flip the prediction of our model (computed using the recently proposed fast minimum norm computation method \cite{pintor2021fast}) is higher for our model for all the $\ell_p$ attacks considered. 


We have also obtained substantial gains in adversarial accuracy against all four $\ell_p$ norm attacks ($p=0,1,2,\infty$) used as benchmarks in adversarial machine learning. \autoref{table:standard_table} displays a subset of results demonstrating accuracy gains against noise and adversarial perturbations, at the expense of a slight decrease in clean accuracy.



\noindent
\noindent
{\bf Ablation:} Since we have different components in our HaH blocks, we explore the effectiveness of each component by doing an ablation study. \autoref{table:ablation_table} summarizes the contribution from each of the components. We see that all of the components (HaH training, divisive normalization, adaptive thresholding) play an important role in obtaining the reported gains in robustness to noise and adversarial attacks. 






\begin{table}[!b]\centering
  \caption{Accuracies for ablation study}
\vspace*{-0.2cm}
  \begin{small}
  \begin{tabular}{@{}rcccc@{}}
    \toprule
    & Clean  &  \makecell[c]{Noisy \\ ($\sigma=0.1$)} &  \makecell[c]{Adv ($\ell_\infty$) \\ ($\epsilon=2/255$)} &  \makecell[c]{Adv ($\ell_2$)\\ ($\epsilon=0.25$)}  \\  \midrule
    All included & 87.3\% & \textbf{64.0\%} & \textbf{21.5\%} & \textbf{27.6\%} \\ \midrule
    No HaH loss & 89.7\%& 50.4\% & 8.8\% & 11.7\% \\  \midrule
    \makecell[r]{Batch norm \\ instead of \\ divisive norm}  & \textbf{90.4\%} & 46.7\%& 12.3\% & 17.4\% \\  \midrule
     \makecell[r]{No \\ thresholding } & 89.9\% & 37.5\% & 3.7\% & 2.5\% \\ 
    \bottomrule
  \end{tabular}
  \end{small}
  \label{table:ablation_table}
\end{table}

\vspace*{-0.15cm}
\section{Conclusion}
\vspace*{-0.15cm}
\label{sec:conc}

Our results indicate the promise of incorporating appropriately engineered neuro-inspired principles into DNN architectures and training. We have chosen supervised learning without augmentation for this initial exposition, but we hope these results motivate further exploration in developing a fundamental understanding of HaH training and inference, as well as in extensive experimentation with a variety of architectures, training techniques (including unsupervised and semi-supervised learning, and data augmentation) and applications.  


\bibliographystyle{IEEEbib}
\bibliography{references}

\end{document}